\documentclass[10pt,twocolumn,letterpaper]{article}

\usepackage{cvpr}              

\usepackage{graphicx}
\usepackage{amsmath}
\usepackage{amssymb}
\usepackage{booktabs}

\usepackage{multirow}
\usepackage{tabularx}
\usepackage{array}
\usepackage{xcolor}
\usepackage{amsmath}

\usepackage[accsupp]{axessibility}  

\usepackage[pagebackref,breaklinks,colorlinks]{hyperref}

\newcommand{\PreserveBackslash}[1]{\let\temp=\\#1\let\\=\temp}
\newcolumntype{C}[1]{>{\PreserveBackslash\centering}p{#1}}
\newcolumntype{R}[1]{>{\PreserveBackslash\raggedleft}p{#1}}
\newcolumntype{L}[1]{>{\PreserveBackslash\raggedright}p{#1}}

\usepackage[capitalize]{cleveref}
\crefname{section}{Sec.}{Secs.}
\Crefname{section}{Section}{Sections}
\Crefname{table}{Table}{Tables}
\crefname{table}{Tab.}{Tabs.}

\def\cvprPaperID{6133} 
\def\confName{CVPR}
\def\confYear{2022}

\begin{document}

\title{
VISOLO: Grid-Based Space-Time Aggregation for\\ Efficient Online Video Instance Segmentation
\vspace{-2mm}
}

\author{
Su Ho Han$^1$, Sukjun Hwang$^1$, Seoung Wug Oh$^2$, Yeonchool Park$^3$,\vspace{1mm}\\Hyunwoo Kim$^4$, Min-Jung Kim$^5$ and Seon Joo Kim$^1$
\vspace{2mm}\\$^1$Yonsei University\qquad$^2$Adobe Research\qquad$^3$LG Electronics\qquad$^4$LG AI Research\qquad$^5$KAIST
\vspace{1mm}\\{\tt\small\{hansuho123, sj.hwang, seonjookim\}@yonsei.ac.kr \qquad seoh@adobe.com} \\{\tt\small yeonchool.park@lge.com \qquad hwkim@lgresearch.ai \qquad emjay73@kaist.ac.kr}
\vspace{-2mm}
}

\maketitle

\begin{abstract}
For online video instance segmentation (VIS), fully utilizing the information from previous frames in an efficient manner is essential for real-time applications.
Most previous methods follow a two-stage approach requiring additional computations such as RPN and RoIAlign, and do not fully exploit the available information in the video for all subtasks in VIS.
In this paper, we propose a novel single-stage framework for online VIS built based on the grid structured feature representation.
The grid-based features allow us to employ fully convolutional networks for real-time processing, and also to easily reuse and share features within different components. 
We also introduce cooperatively operating modules that aggregate information from available frames, in order to enrich the features for all subtasks in VIS. 
Our design fully takes advantage of previous information in a grid form for all tasks in VIS in an efficient way, and we achieved the new state-of-the-art accuracy (38.6 AP and 36.9 AP) and speed (40.0 FPS) on  YouTube-VIS 2019 and 2021 datasets among online VIS methods.
The code is available at \url{https://github.com/SuHoHan95/VISOLO}.
\end{abstract}

\section{Introduction}
Video instance segmentation, introduced in~\cite{Yang_2019_ICCV}, extends instance segmentation in the image domain to the video domain by adding instance tracking. Given a video, all objects in the video need to be located and classified, generating spatio-temporal pixel masks for all objects. VIS is attracting a lot of attention as it is an essential technique for holistic video understanding with various applications such as video editing, autonomous navigation of robots and cars, and augmented reality.

\begin{figure}
\centering
\includegraphics[width=1.0\linewidth]{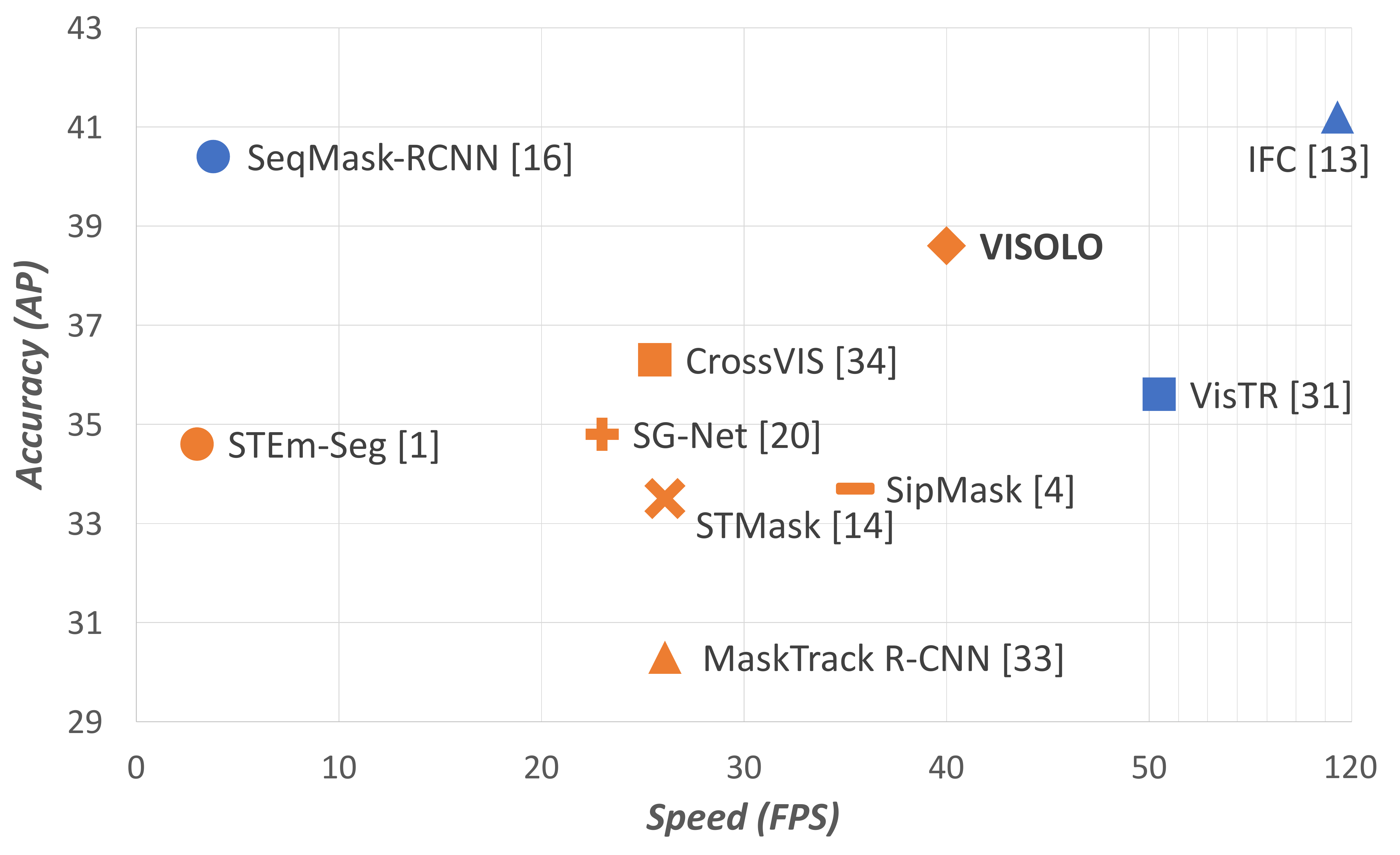}
\vspace{-6mm}
\caption{
Comparisons of the quality and the speed of previous video instance segmentation methods on the YouTube-VIS 2019 dataset. The orange and the blue markers indicate online and offline methods, respectively.
Our framework (VISOLO) is the fastest and the most accurate among online methods, and is approaching the performance of offline methods.
}
\vspace{-4mm}
\label{Fig:intro}
\end{figure}

Recently introduced offline methods for VIS solve the problem through mask propagation and transformers~\cite{Bertasius_2020_CVPR, Lin_2021_ICCV, Wang_2021_CVPR, IFC}.
Although these methods show good performance, they cannot be used for real-time applications as it operates in an offline manner requiring the entire video to be processed before making predictions.

In this paper, we tackle the online video instance segmentation problem where video frames are processed sequentially.
While processing videos online is beneficial for many VIS applications, e.g. robot navigation, it is more challenging compared to the offline approaches as the changes in object appearance over time and the occlusions caused by multiple objects should be handled without information from the future frames.

For online VIS, many algorithms perform frame-level classification and segmentation without fully utilizing the information of previous frames.
However, the inconsistency of object categories and the errors in object masks degrade the performance of VIS when tracking instances from frame-level results. 
For example, MaskTrack R-CNN~\cite{Yang_2019_ICCV} uses the classification and segmentation results of Mask R-CNN~\cite{He_2017_ICCV}, and only uses the previous frame information for tracking.
SipMask~\cite{Cao_SipMask_ECCV_2020} and SG-Net~\cite{liu2021sg} improve the image-level segmentation performance without using the temporal cues that are available from previous frames.
Although CrossVIS~\cite{Yang_2021_ICCV} uses temporal information to enhance the instance features for segmentation and tracking during the training, it does not use temporal information during the inference and for classification.
STMask~\cite{STMask-CVPR2021} uses the information from the previous frame, but it only uses an adjacent frame for segmentation.

In this paper, we propose a framework that exploits information from previous frames not only for tracking but also for classification and segmentation, which is beneficial for increasing the overall VIS performance.
The motivation of our new design is that fully utilizing the available information from past frames for all subtasks is important, as online VIS cannot access the whole video like offline VIS.

In addition, real-time processing is important for online VIS applications. 
Therefore, it is necessary to come up with a framework that can fully utilize the information from previous frames while running real-time. 
CompFeat~\cite{fu2021compfeat} recently proposed  temporal and spatial attention modules that aggregate temporal features for segmentation and classification with non-local operation~\cite{Wang_2018_CVPR}.
However, it requires heavy computation as it employs the two-stage framework based on Mask-RCNN and includes additional encoder to obtain \textit{key} and \textit{value} features as in STM~\cite{Oh_2019_ICCV}.

To this end,  we introduce a novel real-time video instance segmentation framework called VISOLO.
As the name suggests, our work builds upon recently introduced single-stage image instance segmentation SOLO~\cite{wang2020solo, wang2020solov2}, which divides input image into uniform grids and outputs per-grid semantic category scores and instance masks.
Using the grid representation for single-stage VIS has several advantages. 
It enhances the speed by employing a fully convolutional network structure and getting rid of intermediate stages like RPN~\cite{renNIPS15fasterrcnn} and RoI-Align~\cite{He_2017_ICCV}.
It also becomes easier to manage and store features in the grid structure, enabling the addition of extra modules to share the features for multiple subtasks (classification, segmentation and tracking), which in turn upgrades the overall VIS performance.

To fully take advantage of the grid-structured representation, we add a memory matching module that computes the similarity between grids of different frames.
The computed grid similarity is then used for instance matching. 
By storing the grid-structured feature maps of previous frames in memory, the grid similarity can be computed at any time through the memory matching module. This enables us to gain robustness against occlusions and reappearance. 

Furthermore, we propose additional modules called the temporal aggregation and the score reweighting modules, which utilize the information from the previous frames to enhance both the classification and the segmentation performance by using the stored feature maps with a marginal overhead. In VISOLO, subtask heads (classification, segmentation and tracking)  operate interdependently using grid-structured features, so they can  share features effectively and be optimized as a whole network.

Technical contributions of our work are as follows:
\begin{itemize}
    \item We propose a novel online video instance segmentation framework built upon the grid structured representation of SOLO~\cite{wang2020solo,wang2020solov2}. 
    With the grid structure, we can build a single-stage VIS algorithm and avoid computation heavy processes like RPN and RoIAlign in two-stage methods, achieving real-time performance. 
    \item 
    We introduce novel modules -- memory matching, temporal aggregation, and score reweighting modules,
    all of which take advantage of easily manageable grid structured features.
    With the memory that stores features of previous frames, these modules work collaboratively to enrich the features for each subtask of VIS, resulting in the overall VIS performance gain. 
    \item 
    We achieve the new state-of-the-art accuracy on the YouTube-VIS 2019 and 2021 datasets~\cite{Yang_2019_ICCV} (38.6 AP and 36.9 AP) compared to all other online VIS methods.
    Our method also runs in real-time (40.0 FPS), which is the fastest among online algorithms (Fig.~\ref{Fig:intro}).
\end{itemize}

\section{Related Work}
\subsection{Image Instance Segmentation}
Instance segmentation is a task of classifying every pixel in an image to an object category and grouping them into object instances, and serve as the base for VIS.
Instance segmentation can be divided into two-stage~\cite{He_2017_ICCV, Fang_2019_ICCV, Chen_2019_CVPR, Huang_2019_CVPR, Liu_2018_CVPR, Li_2017_CVPR} and single-stage~\cite{wang2020solo, wang2020solov2, Bolya_2019_ICCV, Peng_2020_CVPR, Xu_2019_ICCV, Chen_2020_CVPR} methods.
Two-stage approaches first generate object proposals using Region Proposal Network (RPN)~\cite{renNIPS15fasterrcnn}, then perform box regression, classification and mask prediction using aggregated RoI features.
Single-stage approaches do not use the proposal generation and  employ a fully convolutional network structure to directly predict the bounding-boxes and instance masks.

Most VIS algorithms extend image instance segmentation by adding a tracking head.
Many previous VIS methods~\cite{Bertasius_2020_CVPR, Lin_2021_ICCV, fu2021compfeat, Yang_2019_ICCV} adopt Mask R-CNN~\cite{He_2017_ICCV}, a two-stage approach, and track instances across the video by adding a tracking head or a mask propagation head.
In contrast, our method adopts a single-stage approach with grid-structured representation~\cite{wang2020solo}, which allows us to employ fully convolutional network
to gain efficiency.

\begin{figure*}
\centering
\includegraphics[width=1.0\linewidth]{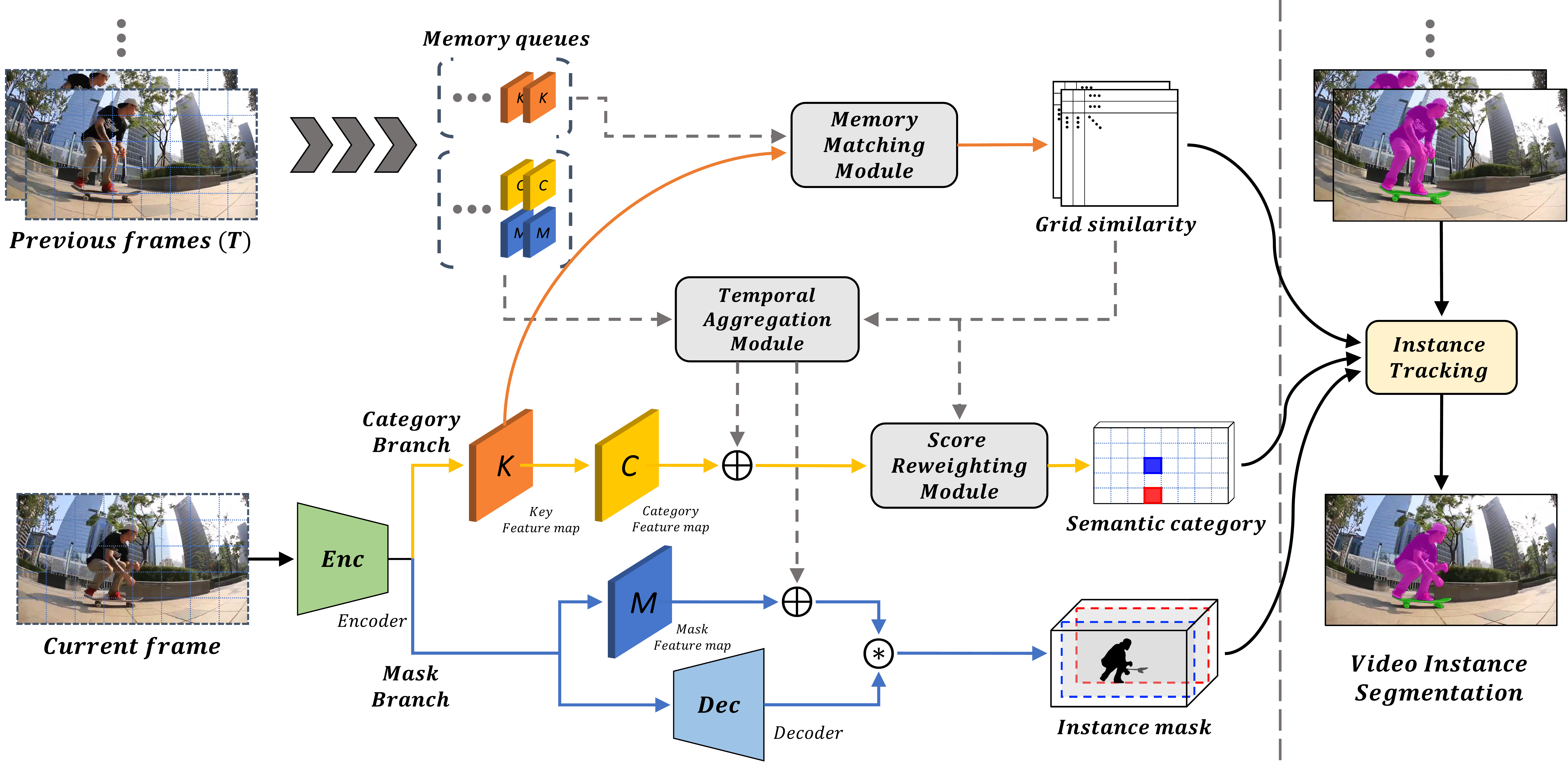}
\vspace{-3mm}
\caption{
Overview of our framework VISOLO.
We take ResNet50~\cite{He_2016_CVPR} as the backbone network for the encoder.
Our network consists of two branches: category branch, mask branch with three additional modules.
The key (K), category (C) and mask(M) feature maps from the category and the mask branch are stored in the memory queues for future use.
Dot arrows denote the use of the information from previous frames.
'$\oplus$' denotes element-wise summation and '$\circledast$' denotes convolution.
}
\vspace{-2mm}
\label{Fig:networks}
\end{figure*}

\subsection{Video Instance Segmentation}
\paragraph{Offline Approaches.}
Many recently introduced methods for VIS solve the problem in an offline manner through mask propagation and transformers~\cite{Bertasius_2020_CVPR, Lin_2021_ICCV, Wang_2021_CVPR, IFC}.
MaskProp~\cite{Bertasius_2020_CVPR} generates multiple overlapping clips by propagating instance masks.
Then the clip-level tracks are aggregated to create instance sequences for the entire video.
Unlike MaskProp\cite{Bertasius_2020_CVPR}, SeqMask R-CNN~\cite{Lin_2021_ICCV} generates instance sequences for the entire video by propagating the instance masks from multiple key frames and reduces redundant sequences.
Recently, transformer based VIS systems have also been introduced (VisTR~\cite{Wang_2021_CVPR} and IFC~\cite{IFC}), which extends DETR~\cite{carion2020endtoend} to the VIS task.

\paragraph{Online Approaches.}
Although the aforementioned methods have shown good performance for VIS, they are restricted in real-time applications as they operate offline, requiring the entire video to be processed before the predictions. Recently, many methods for tackling the online video instance segmentation task have been introduced~\cite{Yang_2019_ICCV, Cao_SipMask_ECCV_2020, liu2021sg, Yang_2021_ICCV, STMask-CVPR2021, fu2021compfeat}.

MaskTrack R-CNN~\cite{Yang_2019_ICCV} is built based on Mask R-CNN~\cite{He_2017_ICCV} by adding a tracking branch that assigns an instance label to each candidate box.
SipMask~\cite{Cao_SipMask_ECCV_2020} proposes a spatial preservation (SP) module to improve the mask prediction performance.
It also adds a branch that generates the tracking feature maps.
Then it matches the instances of frames using these feature maps with a similar metric to MaskTrack R-CNN~\cite{Yang_2019_ICCV}.
SG-Net~\cite{liu2021sg} improves the mask prediction performance by dynamically dividing a target instance into sub-regions and performing segmentation on each regions and add tracking head, which tracks the center of instances.
CrossVIS~\cite{Yang_2021_ICCV} proposes a crossover learning scheme that uses the instance feature in the current frame to pixel-wisely localize the same instance in other frames.
STMask~\cite{STMask-CVPR2021} proposes a spatial calibration to obtain more precise spatial features of anchor boxes. It also adds a temporal fusion module that obtains temporal correlation between adjacent frames to infer instance masks and tracking.
CompFeat~\cite{fu2021compfeat} proposes the temporal and spatial attention module, which aggregate the temporal features to obtain  segmentation and classification results with non-local matching~\cite{Wang_2018_CVPR}.
It also proposes the correlation-based tracking module, which generates both spatial likelihood and object similarity for tracking.

While online approaches show promising results, these methods do not make full use of the information of previous frames.
With a novel design that efficiently stores, manages, and aggregates grid-based features in space and time, VISOLO runs real-time, and improves the performance by a good margin compared to previous online methods, approaching the accuracy of the offline approaches.


\section{Method Overview}
The overall framework of VISOLO is illustrated in Fig.~\ref{Fig:networks}.
With the grid-structured feature representation from SOLO~\cite{wang2020solo} (Mask and Category Branch) and the memory that stores grid features from previous frames, we add three modules that effectively aggregate features for each subtask of VIS.

\subsection{SOLO Review}
SOLO is a recently introduced and high-performing method for single-stage image instance segmentation.
In SOLO, an input image is first conceptually divided into uniform grids ${S\times S}$, which then goes through two branches: category branch and mask branch.
If a grid cell contains the center of an object, that grid cell is responsible for predicting the semantic category and the instance mask of that object in the category branch and mask branch respectively.
In SOLOv2~\cite{wang2020solov2}, authors introduced the dynamic head, in which the mask branch is decoupled into a feature branch and a kernel branch.
The feature branch predicts the fine-grained instance-aware feature map through decoder, and the kernel branch predicts ${1\times 1}$ convolution kernel weights conditioned on the location for each grid.
The mask branch outputs an instance mask for each grid by running convolution on the feature map with the generated kernel weights.
We use this dynamic head as the mask branch.

We made several structural modifications on the mask and the category branches to make them fit to VIS.
First, our mask branch no longer relies on feature pyramid networks (FPN)~\cite{Lin_2017_CVPR}.
FPN is not effective for the current VIS benchmark~\cite{Yang_2019_ICCV} that mostly consists of large and distinct objects.
Also, its multi-level nature brings many difficulties during optimization of the grid similarities.
Instead, we employ encoder-decoder structure that has shown to be effective for generating high-quality object masks in video segmentation~\cite{Oh_2018_CVPR, Oh_2020_PAMI}. 
Second, the intermediate features -- key feature map (K), category feature map (C) and mask feature map (M) in Fig.~\ref{Fig:networks} -- of the category and the mask branch are used to take the richer temporal information.
Specifically, the key feature map (K) from the category branch is used to obtain the grid similarities with the previous frames through the memory matching module.
The category (C) and the mask feature maps (M) from the category and the mask branches are enhanced through the temporal aggregation module, in order to improve per-frame classification and segmentation.

\subsection{VISOLO Overview}
VISOLO is designed to utilize grid shaped features of SOLO for VIS, by enriching the features through temporal aggregation. 
For each frame, the computed features (K, C, M) are stored in the memory queues. 
The memory matching module predicts the grid similarity by comparing the Key features (K) between frames.
The computed grid similarity is used for tracking the instances across frames as well as gathering the information from previous frames.
The temporal aggregation module is responsible for providing the category branch and the mask branch with rich information of previous frames.
In this module, Category features (C) and Mask features (M) of previous frames in the memory are combined using the grid similarity. 
The score reweighting module is used to enhance the classification in the category branch by dynamically calibrating the output score map.
Note that the score reweighting module also uses the computed grid similarity. 

The effective design of feature aggregation and flow is what separates our method from previous approaches. 
Previous online methods did not make a full use of information from previous frames for all subtasks in VIS. 
In contrast, the components and the modules in our framework are tightly coupled
with efficient flow of information between them.
The information flow from the previous frames to the current frame adds only a marginal overhead as they exploit already computed grid-structured features and grid similarity from the memory matching module.
This allows VISOLO to maximize the use of available information, resulting in a high performance VIS system that runs real-time.

\begin{figure}
\centering
\includegraphics[width=1.0\linewidth]{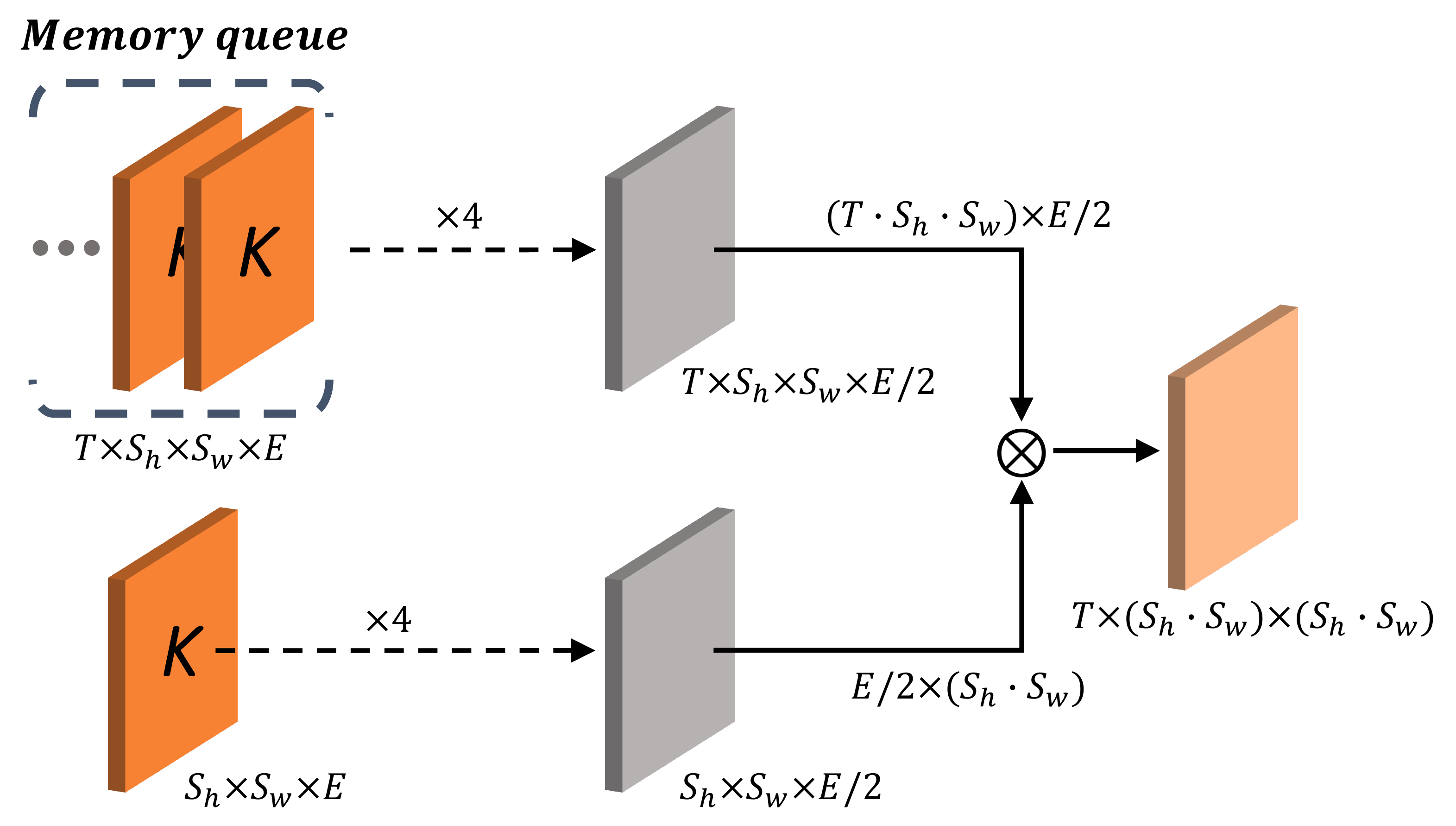}
\caption{
Detailed implementation of the memory matching module operation. It takes the key feature maps from the memory queue and the category branch as inputs.
$S_h$ and $S_w$ are the number of grids in height and width, respectively, and $E$ is the input feature map dimension. Dot arrows denote convolutional layers and '$\otimes$' denotes matrix inner-product.
}
\label{Fig:MM_module}
\end{figure}

\section{Method Details}
\subsection{Memory Matching Module}
\label{sect:MM_module}
The memory matching module predicts the grid similarity ($\mathbf{Sim}$) between grids of the current frame and previous frames in memory by iterating pairwise comparison. 
Outputs (key feature maps) from the first layer in the category branch for the current and the previous frames are used as inputs.
The two inputs go through convolutional layers separately and then are merged by the matrix inner-product to generate the grid similarities.
The details for the memory matching module are illustrated in Fig~\ref{Fig:MM_module}.

Because our method outputs the semantic category and the mask of instances for each grid, if we can figure out the grids of previous frames that are identical to the grids corresponding to the instances of the current frame, tracking can be performed by matching instances of those grids. 
Therefore, to obtain the grid information of same instances across frames, we design the memory matching module to predict the grid similarity. Furthermore, the grid similarity is used for reading the instance appearance information of previous frames to enhance the classification and the segmentation performance in the temporal aggregation and the score reweighting modules.

\subsection{Temporal Aggregation Module}
\label{sect:TA_module}
The temporal aggregation module is designed to enhance the classification and the segmentation performance by aggregating temporal information. 
For each grid, the temporal aggregation module gathers the appearance information from the past using the grid similarity from the memory matching module. 
The feature maps of the previous frames in the memory queue ($\mathbf{C}_T \& \mathbf{M}_T\in\mathbb{R}^{T \times S_h \times S_w \times E}$) are further processed by a convolutional layer and reshaped to $\mathbb{R}^{(T \cdot S_h \cdot S_w) \times E}$, then these features are aggregated through a weighted summation, where the weights ($\mathbf{W}_T\in\mathbb{R}^{(S_h \cdot S_w) \times (T \cdot S_h \cdot S_w)}$) are computed by applying softmax function on the grid similarity. 

Specifically, the output feature maps of the Temporal Aggregation Module can be put as follows:
\begin{equation}
    \mathbf{C}_A = \mathbf{W}_T \otimes \mathbf{C}_T,
\end{equation}
\begin{equation}
    \mathbf{M}_A = \mathbf{W}_T \otimes \mathbf{M}_T,
\end{equation}
where '$\otimes$' denotes matrix inner-product. The retrieved features $\mathbf{C}_A$ and $\mathbf{M}_A$ are added with category and mask feature maps of the current frame.

Our temporal aggregation operation is related to the operation in STM~\cite{Oh_2019_ICCV, Oh_2020_PAMI} in that both retrieve the appearance information from previous frames with soft weights.
However, STM~\cite{Oh_2019_ICCV, Oh_2020_PAMI} needs to re-encode every additional memory frame through ResNet encoder to obtain the \textit{value} feature.  
On the other hand, our temporal aggregation module works more efficiently by reusing the category and the mask branch's features as the \textit{value}.

\begin{figure}
\centering
\includegraphics[width=1.0\linewidth]{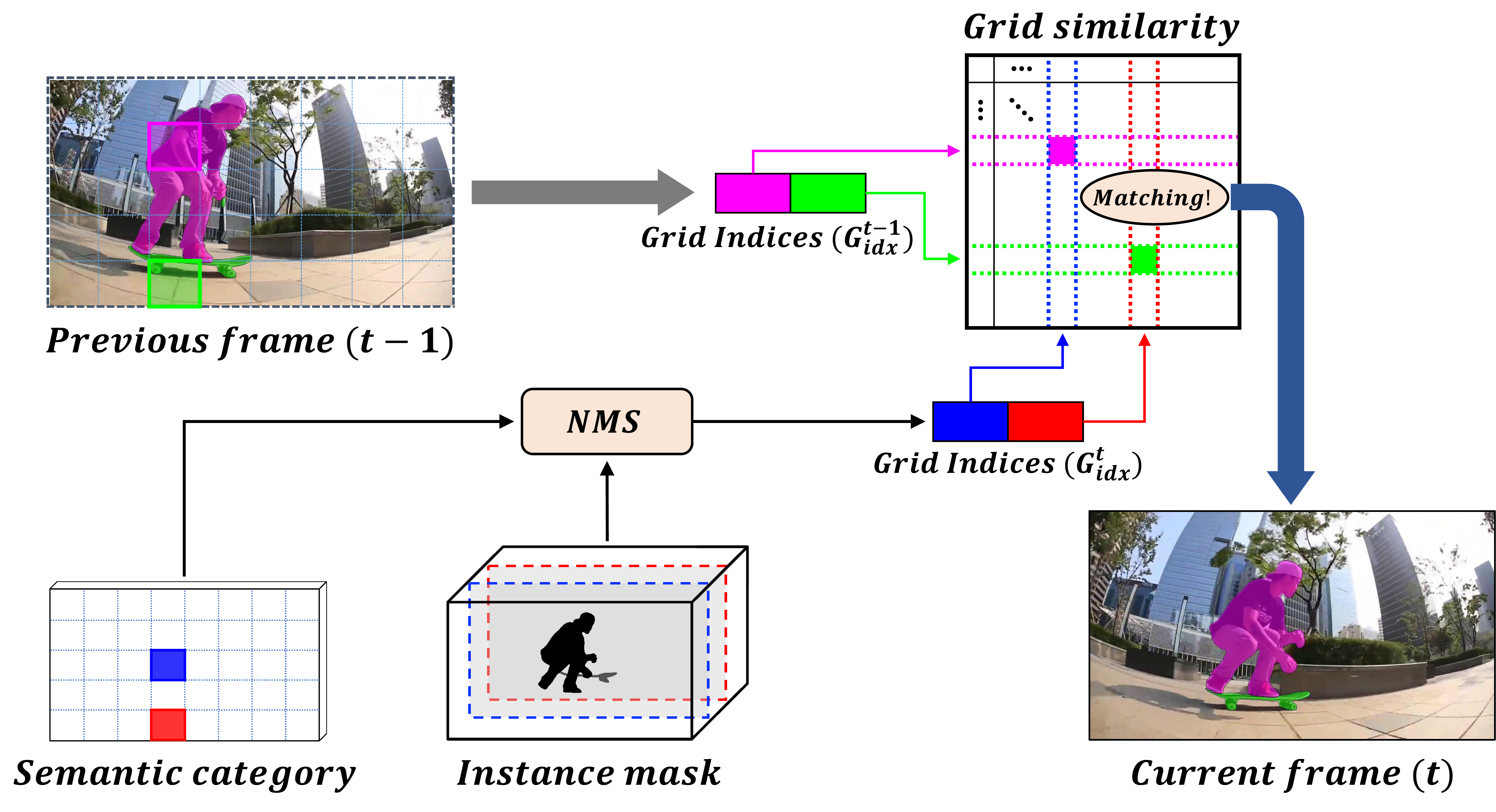}
\caption{
Overview of instance tracking operation between the current frame ($t$) and the previous frame (${t-1}$).
$G_{idx}$ indicates the index of grids that contain the center of the instances.
}
\label{Fig:Tracking}
\end{figure}

\subsection{Score Reweighting Module}
\label{sect:SR_module}
To further enhance the classification performance, we propose the score reweighting module, which  dynamically calibrates the output score from the category branch.
Because the score reweighting module works using basic tensor operations with the grid similarity already computed in the memory matching module, it adds only a small overhead.

The score reweighting module computes the weights for each grid in $\mathbf{Cat}\in\mathbb{R}^{S_h \times S_w \times C}$, which is the output score from the category branch, using the grid similarities from the memory matching module. 
We use two grid similarities between the current frame and two previous frames ($\mathbf{Sim}_1$ and $\mathbf{Sim}_2$).
To match the dimension of $\mathbf{Cat}$, each similarity matrix $(\mathbf{Sim}\in\mathbb{R}^{(S_h \cdot S_w) \times (S_h \cdot S_w)})$ is reshaped first by taking the maximum of each row of the similarity matrix and then converting it to a matrix of the size ${S_h \times S_w}$.
For convenience, we will call the reshaped similarities as $\mathbf{\Tilde{Sim}}_1$ and $\mathbf{\Tilde{Sim}}_2$.
Each grid in $\mathbf{Cat}$ is multiplied with the average of the corresponding grid in $\mathbf{\Tilde{Sim}}_1$ and $\mathbf{\Tilde{Sim}}_2$, resulting in the final classification score $\mathbf{P}$:
\begin{equation}
    \mathbf{P} = \mathbf{Cat} \circledcirc \textsc{avg}(\mathbf{\Tilde{Sim}}_1, \mathbf{\Tilde{Sim}}_2),
\end{equation}
where $\circledcirc$ denotes element-wise multiplication for each channel.

\begin{table*}
\begin{center}
\begin{tabularx}{0.9\linewidth}{@{\extracolsep{\fill}} c|c|c|c|ccccc}
\toprule
\multicolumn{2}{c|}{Methods} & Backbone & FPS & AP & AP$_{50}$ & AP$_{75}$ & AR$_1$ & AR$_{10}$\\
\midrule
\multirow{3}*{Offline} & MaskProp~\cite{Bertasius_2020_CVPR} & ResNet-50 & $-$ & 40.0 & $-$ & 42.9 & $-$ & $-$\\
& SeqMask-RCNN~\cite{Lin_2021_ICCV} & ResNet-50 & 3.8 & 40.4 & 63.0 & 43.8 & 41.1 & 49.7\\
& VisTR~\cite{Wang_2021_CVPR} & ResNet-50 & 51.1 & 35.6 & 56.8 & 37.0 & 35.2 & 40.2\\
& IFC~\cite{IFC} & ResNet-50 & 107.1 & 41.2 & 65.1 & 44.6 & 42.3 & 49.6\\
\midrule
Near Online & STEm-Seg~\cite{Athar_Mahadevan20ECCV} & ResNet-101 & 3.0 & 34.6 & 55.8 & 37.9 & 34.4 & 41.6\\
\midrule
\multirow{10}*{Online} & MaskTrack-RCNN~\cite{Yang_2019_ICCV} & ResNet-50 & 26.1 & 30.3 & 51.1 & 32.6 & 31.0 & 35.5\\
& SipMask~\cite{Cao_SipMask_ECCV_2020} & ResNet-50 & 35.5 & 33.7 & 54.1 & 35.8 & 35.4 & 40.1\\
& SG-Net~\cite{liu2021sg} & ResNet-50 & 23.0$^{\ast}$ & 34.8 & 56.1 & 36.8 & 35.8 & 40.8\\
& SG-Net~\cite{liu2021sg} & ResNet-101 & 19.8$^{\ast}$ & 36.3 & 57.1 & 39.6 & 35.9 & 43.0\\
& CompFeat~\cite{fu2021compfeat} & ResNet-50 & $-$ & 35.3 & 56.0 & 38.6 & 33.1 & 40.3\\
& CrossVIS~\cite{Yang_2021_ICCV} & ResNet-50 & 25.6 & 36.3 & 56.8 & 38.9 & 35.6 & 40.7\\
& CrossVIS~\cite{Yang_2021_ICCV} & ResNet-101 & 23.3 & 36.6 & 57.3 & 39.7 & 36.0 & 42.0\\
& STMask~\cite{STMask-CVPR2021} & ResNet-50$^{\dagger}$ & 26.1 & 33.5 & 52.1 & 36.9 & 31.1 & 39.2\\
& STMask~\cite{STMask-CVPR2021} & ResNet-101$^{\ddagger}$ & 22.4 & 36.8 & 56.8 & 38.0 & 34.8 & 41.8\\
& \textbf{Our VISOLO} & ResNet-50 & 40.0 & 38.6 & 56.3 & 43.7 & 35.7 & 42.5\\
\bottomrule
\end{tabularx}
\end{center}
\vspace{-2mm}
\caption{Quantitative evaluation on \textbf{YouTube-VIS 2019}~\cite{Yang_2019_ICCV} validation set. \cite{liu2021sg} does not provide official checkpoints, so we infer the speed reported in \cite{liu2021sg} (FPS with superscript "${\ast}$"). "${\dagger}$" and "${\ddagger}$" indicate the ResNet-50-DCN and ResNet-101-DCN, respectively.}
\label{Table:YTVis}
\end{table*}

\subsection{Instance Tracking}
The instance tracking operation is illustrated in Fig.~\ref{Fig:Tracking}.
At current frame $t$, we first apply the Matrix NMS~\cite{wang2020solov2} to the outputs from the mask branch and the category branch to obtain indices of grids $G_{idx}^t$ that contain the center of instances.
The tracking is achieved by comparing the grid similarity values between the $G_{idx}^t$ of the current frame and the $G_{idx}^{t-1}$ of the previous frame.
The current instances are basically connected to the previous instances with the highest similarity value.
If the similarities between the $G_{idx}^t$ and the $G_{idx}^{t-1}$ are all below a certain threshold value (0.1), we then move on to the next previous frame, which contains the instances that failed to track, and use the grid similarity for that past frame to see if the current instance matches with any instances in that frame.
Note that we keep the feature map, which is used to compute grid similarity, and the $G_{idx}^{fail}$ of the previous frame and the past frames that contain the instances that failed to track.
This allows us to handle occlusion and reappearance problems.
If the current instance does not match with any previous instances, it is declared as a new instance.

\subsection{Training and Inference}
We jointly train classification, segmentation and grid similarity prediction tasks in an end-to-end manner. We define our training loss function as follows:
\begin{equation}\label{eqloss}
    \mathbf{L} = \mathbf{L}_{class} + \lambda\mathbf{L}_{mask} + \mathbf{L}_{grid},
\end{equation}
Each loss corresponds to the category branch loss, the mask branch loss, and the grid similarity loss respectively. 
We use the Focal Loss~\cite{Lin_2017_ICCV} for $\mathbf{L}_{class}$ and $\mathbf{L}_{grid}$, and the Dice Loss~\cite{7785132} for $\mathbf{L}_{mask}$. $\mathbf{L}_{mask}$ and $\mathbf{L}_{grid}$ are computed only for grids where the ground truth objects exist.
$\lambda$ in equation (\ref{eqloss}) is set to 3.

During the inference, the intermediate features from the category and the mask branches are stored onto the external memory.
For the temporal aggregation module, using more frames helped in enriching the intermediate features of the category and the mask branches. 
However, writing features of all previous frames onto the memory is inefficient, so we select frames to be kept in the memory by a simple rule.
By default, features of the two previous frames are always saved,
as they provide valuable appearance information for tracking and also are used in the score reweighting module.
For the intermediate frames, we simply save features of every 5 frames.
For the first frame in the input video, we put it onto the memory and use it as a reference.

\begin{table}
\begin{center}
\begin{tabular}{C{2.8cm}|C{0.6cm}C{0.6cm}C{0.6cm}C{0.6cm}C{0.6cm}}
\toprule
Methods & AP & AP$_{50}$ & AP$_{75}$ & AR$_1$ & AR$_{10}$\\
\midrule
MaskTrack-RCNN &  28.6 & 48.9 & 29.6 & 26.5 & 33.8\\
SipMask & 31.7 & 52.5 & 34.0 & 30.8 & 37.8\\
CrossVIS & 34.2 & 54.4 & 37.9 & 30.4 & 38.2\\
STMask & 30.6 & 49.4 & 32.0 & 26.4 & 36.0\\
\textbf{Our VISOLO} & 36.9 & 54.7 & 40.2 & 30.6 & 40.9\\
\bottomrule
\end{tabular}
\end{center}
\vspace{-2mm}
\caption{
Quantitative evaluation on \textbf{YouTube-VIS 2021} validation set. We refer the results reported in \cite{Yang_2021_ICCV}. All models use the ResNet-50~\cite{He_2016_CVPR} as the backbone network, except for STMask~\cite{STMask-CVPR2021} that uses ResNet-50-DCN.
}
\vspace{-2mm}
\label{Table:YTVis2021}
\end{table}

\begin{figure*}
\centering
\includegraphics[width=1.0\linewidth]{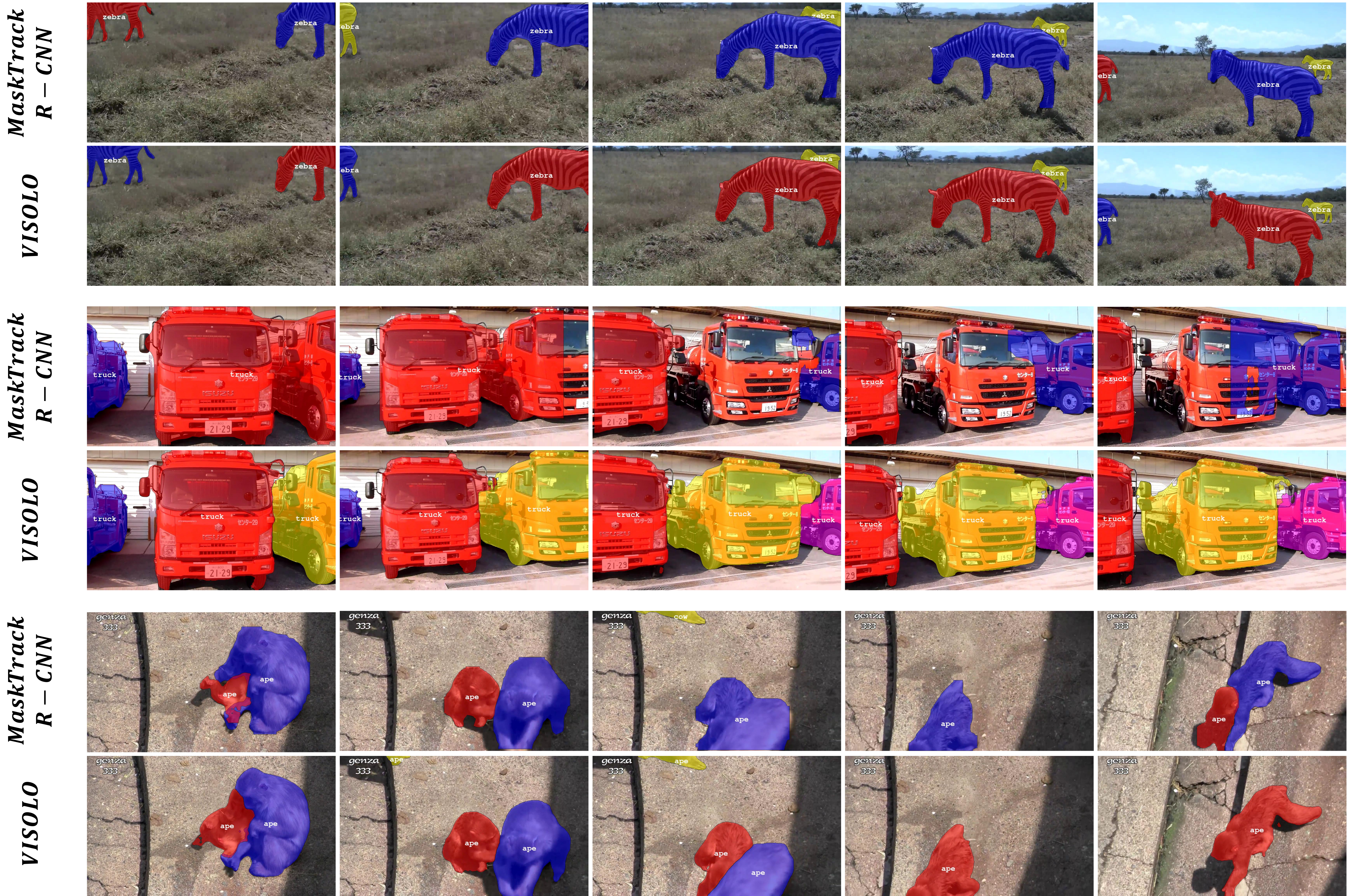}
\caption{
We compare our video instance segmentation results with MaskTrack R-CNN~\cite{Yang_2019_ICCV} results. Different instances are assigned with different colors.
}
\vspace{-2mm}
\label{Fig:QualiResult}
\end{figure*}

\section{Experiments}
We evaluate our model using the ResNet50 backbone on the YouTube-VIS 2019 and 2021~\cite{Yang_2019_ICCV} datasets. YouTube-VIS 2019 is the first large-scale dataset for the video instance segmentation and YouTube-VIS 2021 is an improved version of YouTube-VIS 2019.
We perform our evaluations on the validation sets of the YouTube-VIS 2019 and 2021 dataset, which consist of 302 videos and 421 videos, respectively.
We measured the mean video average precision (AP), video average precision with IoU threshold 50\% and 75\% (AP$_{50}$, AP$_{75}$), average recall given 1 and 10 instances per video (AR$_1$, AR$_{10}$) and frames per second (FPS).

\begin{table}
\begin{center}
\begin{tabular}{C{1.0cm}C{1.4cm}C{1.0cm}|C{0.6cm}C{0.6cm}C{0.6cm}}
\toprule
\multirow{2}*{SR} & TA & TA & \multirow{2}*{AP} & \multirow{2}*{AP$_{50}$} & \multirow{2}*{AP$_{75}$}\\
 & (Category) & (Mask) & & & \\
\midrule
 & & & 34.6 & 51.5 & 36.8\\
 \checkmark & & & 35.6 & 53.8 & 37.9\\
 & \checkmark & & 36.4 & 54.4 & 39.3\\
 \checkmark & \checkmark & & 37.7 & 56.6 & 40.3\\
 \checkmark & \checkmark & \checkmark & 38.6 & 56.3 & 43.7\\
\bottomrule
\end{tabular}
\end{center}
\vspace{-4mm}
\caption{Ablation study of the \textbf{S}core \textbf{R}eweighting module (SR) and the \textbf{T}emporal \textbf{A}ggregation module (TA), estimated on YouTube-VIS 2019 dataset.}
\vspace{-4mm}
\label{Table:modules}
\end{table}

\subsection{Implementation Detail}
For the training, we use the image instance segmentation dataset COCO~\cite{DBLP:conf/eccv/LinMBHPRDZ14}, and two video instance segmentation datasets YouTube-VIS 2019 and 2021~\cite{Yang_2019_ICCV}.
To exploit the static image dataset for video instance segmentation, we transformed static images into  3-frame synthetic videos using random affine transformations.

\begin{table}
\begin{center}
\begin{tabular}{L{2.5cm}|C{0.8cm}|C{0.8cm}C{0.8cm}C{0.8cm}}
\toprule
Memory frames & FPS & AP & AP$_{50}$ & AP$_{75}$\\
\midrule
2 frames & 40.4 & 36.7 & 54.2 & 40.4\\
10 frames & 39.5 & 37.5 & 55.3 & 41.3\\
20 frames & 38.7 & 37.7 & 55.4 & 41.4\\
Every 5 frames & 40.0 & 38.6 & 56.3 & 43.7 \\
\bottomrule
\end{tabular}
\end{center}
\vspace{-4mm}
\caption{The number of reference frames for temporal aggregation module analysis on the validation sets of YouTube-VIS 2019 dataset~\cite{Yang_2019_ICCV}. We compare results by different memory storing rules.}
\vspace{-4mm}
\label{Table:Number}
\end{table}

First, our network is trained on the COCO~\cite{DBLP:conf/eccv/LinMBHPRDZ14} dataset for pre-training with the batch size 16 for 60 epochs and the initial learning rate of 1e-4, which decays at 40 epochs.
After pre-training, we fine-tune the our network on the youtube VIS 2019 and 2021 datasets together with COCO dataset to prevent overfitting. When combining datasets, we use 21 classes of COCO that are related with 40 classes of YouTube-VIS 2019 and 2021, respectively. During fine-tuning, we sampled training data with the following distribution: (YouTube-VIS 2019 (75\%), COCO (25\%)) and (YouTube-VIS 2021 (75\%), COCO (25\%)), depending on the test dataset. In fine-tuning, our network trained with batch size 20 for 68 epochs iterations and the initial learning rate of 1e-4, which decays at 30 epochs and 52 epochs. In both training stage, each batch consists of 3 frames.

\begin{figure*}
\centering
\includegraphics[width=1.0\linewidth]{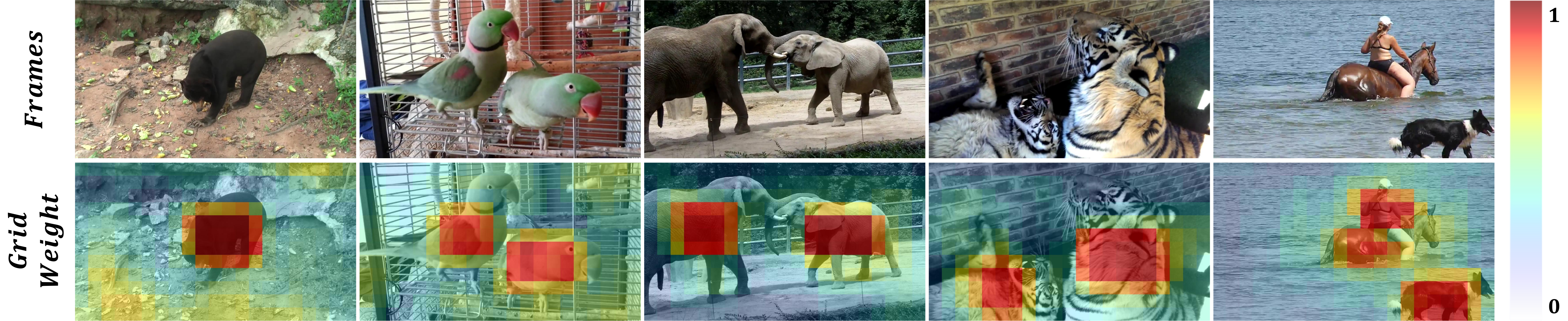}
\vspace{-6mm}
\caption{
Visualization of weights for each grid in the score reweighting module at the second row. The first row shows the original frames.
}
\vspace{-2mm}
\label{Fig:Score}
\end{figure*}

\begin{figure*}
\centering
\includegraphics[width=1.0\linewidth]{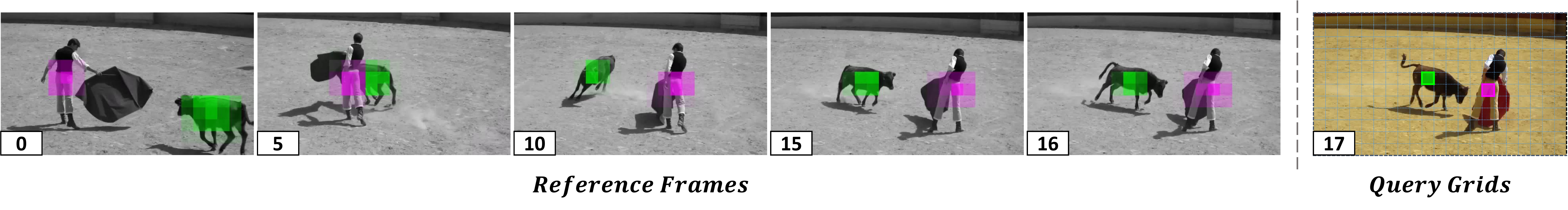}
\vspace{-6mm}
\caption{
Visualization of our temporal aggregation module operation. We first compute the grid similarities between query grids and all grids of reference frames, and obtain the soft weight by a softmax operation.
Then, we visualize the normalized soft weights of the reference frames. The query grids and weights of each grid of reference frames with respect to the query grids are assigned with different colors.
}
\label{Fig:Memory}
\vspace{-2mm}
\end{figure*}

\begin{figure}
\centering
\includegraphics[width=1.0\linewidth]{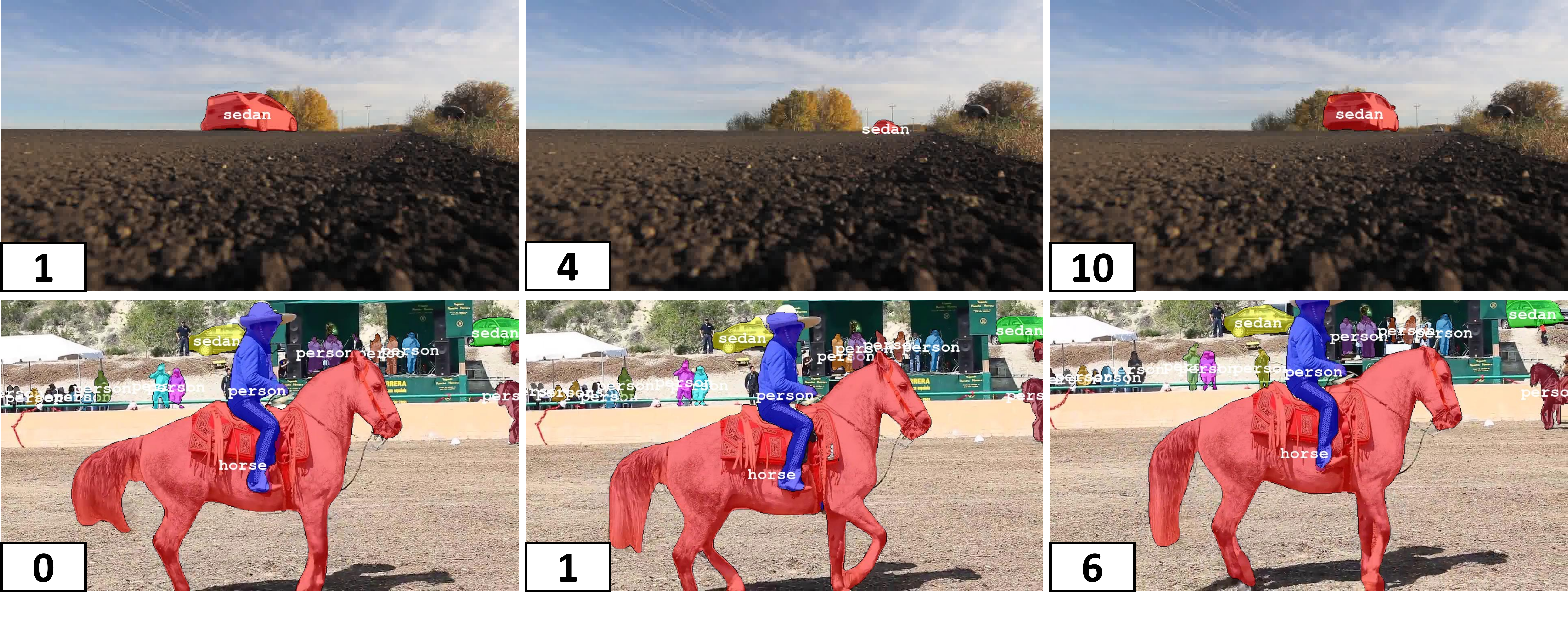}
\vspace{-6mm}
\caption{
Some failure cases of our method.
}
\vspace{-2mm}
\label{Fig:Failure}
\end{figure}

We use randomly cropped 356x624 patches for the training, and the inference image size is set to 356x624, which is the same size as the training patch.
We set the number of grids as the aspect ratio of the input video ($S=(12,21)$).
As for the label assignment, to generate the target category probability and the mask, we use the same metric of SOLO~\cite{wang2020solo}.
For the target grid similarity $\mathbf{Sim}\in\mathbb{R}^{(S_h \cdot S_w) \times (S_h \cdot S_w)}$, we assign '1' to every grid that contain the center region of the same instances between two frames. Otherwise, we assign '0'.

\subsection{Quantitative Results}
\noindent{\bf YouTube-VIS 2019 evaluation results.}
We compare our method with the state-of-the-art offline and online methods in Table~\ref{Table:YTVis}. 
We report both the accuracy and the speed. 
We measure the frames per second (FPS) of most previous methods using the official codes in the same environment with a single 2080 Ti GPU.

As shown in Table~\ref{Table:YTVis}, our method outperforms all the online methods on both accuracy (38.6 AP) and speed (40.0 FPS) by a good margin, closing the gap between the online and the offline VIS algorithms.

\noindent{\bf YouTube-VIS 2021 evaluation results.} 
Table~\ref{Table:YTVis2021} summarizes our performance on YouTube-VIS 2021 validation set. 
Again, our method outperforms all the online methods by a large margin.

\subsection{Qualitative Results}
In Fig.~\ref{Fig:QualiResult}, we compare qualitative results of VISOLO with MaskTrack R-CNN~\cite{Yang_2019_ICCV}.
We chose example videos from the validation videos of YouTube-VIS 2019 dataset~\cite{Yang_2019_ICCV}.
As can be seen in the figure, our method performs better in all aspects of VIS including instance classification, segmentation, and tracking, even in difficult cases with occlusion and complex motion.

\subsection{Analysis}

\noindent{\bf Contribution of each component.} 
We first conduct ablation studies to verify the contribution of each component in our framework (Table~\ref{Table:modules}).
In this ablation study, we tested three components -- the score reweighting module and the temporal aggregation for the category and the mask branches.
As can be seen in the results, all components play important roles and the best performance is achieved when we combine all the components.

\noindent{\bf Effect of memory frames.} 
We also analyze the effect of storing different numbers of frames on the memory in Table~\ref{Table:Number}.
As shown in the table, all variations of our model reach the state-of-the-art accuracy compared to other online methods, except using two previous frames. 
We choose the model with every 5 intermediate frames as our final choice, considering the trade-off between the accuracy and the speed.

\noindent{\bf Visualizations.} 
In Fig.~\ref{Fig:Score}, we visualize the weights computed in our score reweighting module for each grid.
As can be observed, grids containing the center of objects are accurately emphasized by the score reweighting module.
We also show the soft weights that are used to retrieve the information of the reference frames in the temporal aggregation module, i.e. weights of each grid of the reference frames with respect to the query grids.
Fig.~\ref{Fig:Memory} visualizes the weights of reference grids contributing to the ox and the bullfighter of the query frame.
It shows that our temporal aggregation module accurately gathers the appearance information from the reference frames.

\noindent{\bf Limitations.}
First, we show some failure cases of our method in Fig.~\ref{Fig:Failure}.
If a new object of similar appearances to a previous instance, our method confuses distinguishing the identities.
Also, our method may have difficulty detecting small objects.
The original SOLO~\cite{wang2020solo, wang2020solov2} framework uses feature pyramid networks (FPN) and sets different numbers of grids for each level to detect objects in various sizes.
However, considering the characteristics of YouTube-VIS, comprising objects of large sizes and distinct movements, VISOLO uses only a single level to improve the efficiency.
Therefore, our method can fail detecting small objects if multiple instances belong to the same grid.

Second, VISOLO performance is highly dependent on the grid similarity. The grid similarity computed by the memory matching module is used not only for tracking instances, but also for the temporal aggregation and score reweighting modules. This design makes our method work efficiently, but at the same time, there is a chance of performance drop if there is a problem with the grid similarity.

\section{Conclusion}
In this paper, we presented a novel online video instance segmentation method named VISOLO built based on the grid structured feature representation. 
We fully take advantage of the grid-based features, making the features to be reused and shared to maximize the use of the information from previous frames. 
With our efficiently designed framework, we gained significant improvements in performance, achieving the state-of-the-art accuracy and speed compared to the other online methods on the YouTube-VIS 2019 and 2021 datasets.

\section*{Acknowledgments}
This research was supported by LG Electronics, and  also by Institute of Information \& communications Technology Planning \& evaluation (IITP) grant funded by the Korea government(MSIT), Artificial Intelligence Graduate School Program, Yonsei University, under Grant 2020-0-01361.

{\small
\bibliographystyle{ieee_fullname}
\bibliography{egbib}
}

\end{document}